\documentclass[runningheads]{llncs}

\usepackage{eccv}
\usepackage{eccvabbrv}
\usepackage{graphicx}
\usepackage{booktabs}
\usepackage[accsupp]{axessibility}
\usepackage[utf8]{inputenc}
\usepackage{amsmath}
\usepackage{amssymb}
\usepackage{mathtools}
\usepackage{url}
\usepackage{multirow}
\usepackage{enumitem}
\usepackage[hidelinks]{hyperref}

\newcommand{\loss}[1]{\mathcal{L}_{\text{#1}}}

\newcommand{\corr}{\textsuperscript{*}}

\begin{document}

\title{MAC-Splat: Multi-Attribute Consistency for High-Fidelity Sparse-View Reconstruction}
\titlerunning{MAC-Splat}

\author{
Jinqian Yang\inst{1} \and
Yichen Wu\inst{2}\corr \and
Wanhua Li\inst{3} \and
Haokun Lin\inst{4} \and
Renzhen Wang\inst{5} \and
Xiangchu Feng\inst{1} \and
Xixi Jia\inst{1}\corr
}

\authorrunning{J.~Yang et al.}

\institute{Xidian University, China \and Harvard University, USA \and Nanyang Technological University, Singapore \and City University of Hong Kong, China \and Xi'an Jiaotong University, China}

\maketitle
\begingroup
\renewcommand{\thefootnote}{*}
\footnotetext{Corresponding authors: Yichen Wu
(\href{mailto:wuyichen.am97@gmail.com}{wuyichen.am97@gmail.com})
and Xixi Jia
(\href{mailto:xxjia@xidian.edu.cn}{xxjia@xidian.edu.cn}).}
\endgroup

\begin{abstract}
Reconstructing high-fidelity 3D scenes from sparse views remains a central problem in generalizable neural rendering. Existing generalizable 3D Gaussian Splatting (3DGS) methods often exhibit geometric artifacts in sparse-view settings, since supervision based solely on 2D photometric losses cannot resolve depth and correspondence ambiguities. To address this issue, we propose MAC-Splat, a training framework built around direct 3D consistency supervision. MAC-Splat builds on the MASt3R geometric backbone and a frozen DINOv3 encoder to obtain semantically informed 2D correspondences, which serve as geometric anchors for 3D supervision. Using these anchors, we define the Multi-Attribute Consistency (MAC) loss. This objective jointly regularizes the 3D attributes of matched Gaussians, including their position, shape, and appearance, by enforcing agreement in a common world coordinate frame. The formulation is robust to outliers and respects the geometry of covariance matrices, which leads to stable training under sparse-view conditions. Experiments on ScanNet++ show that MAC-Splat outperforms strong baselines, with particularly large gains under different overlap regimes. In particular, it improves average PSNR over Splatt3R by more than 4.5 dB, reduces LPIPS, and maintains performance as the camera pose gap increases. These results indicate that a direct, multi-attribute 3D consistency objective, when combined with high-quality correspondences, is effective for addressing the ill-posed sparse-view reconstruction problem.
 

\end{abstract}

\section{Introduction}
\label{sec:intro}

Capturing and rendering photorealistic 3D scenes is a central goal in computer vision~\cite{schoenberger2016sfm, yao2018mvsnet, chen2025survey3dgs, wu2024recent}, with applications ranging from virtual reality~\cite{tancik2022block} to robotics~\cite{sucar2021imap}. Implicit neural representations such as NeRF~\cite{ mildenhall2020nerf, barron2021mip} have demonstrated striking visual quality; however, their training and inference costs remain high, which limits real-time deployment and scalability in practical systems.

\begin{figure}[t!]
    \centering
    \includegraphics[width=1\textwidth]{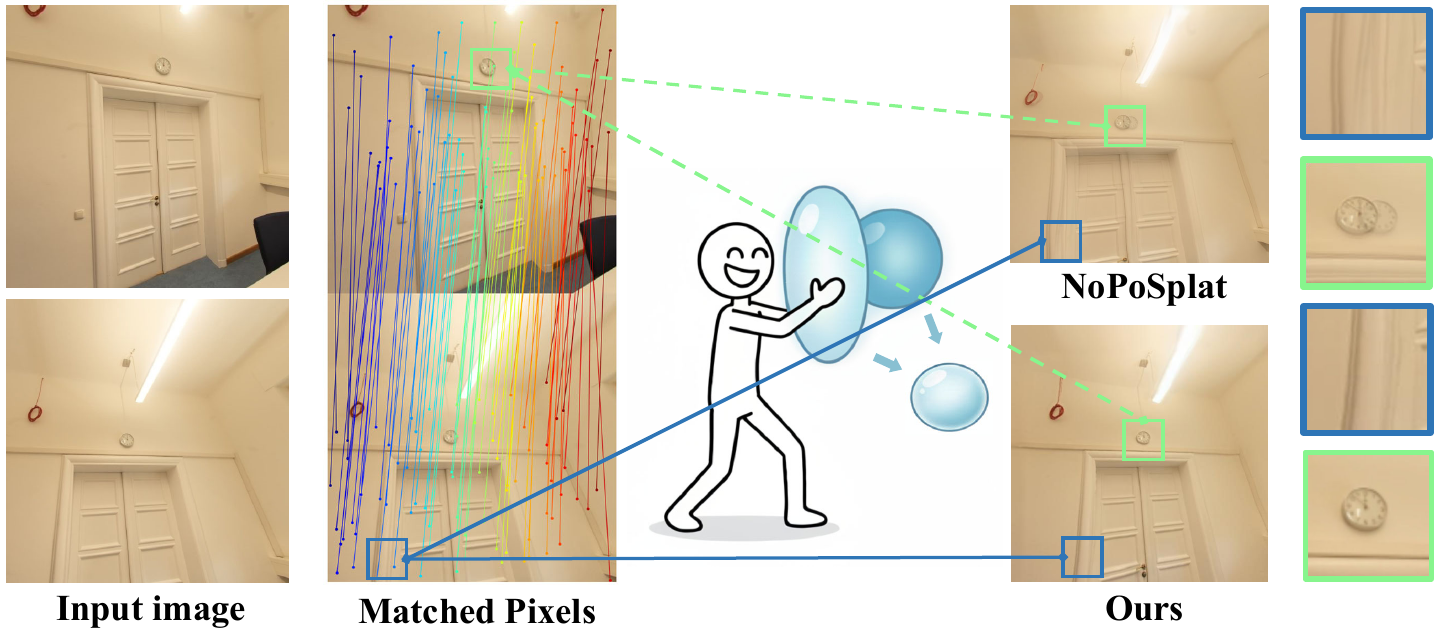}
    \caption{
        \textbf{Matched-pixel constraints for 3D Gaussians.} 
    Pixel correspondences are used as anchors to enforce consistent Gaussian attributes, which helps mitigate common sparse-view artifacts compared to prior generalizable 3DGS methods.
    }
    \label{fig:problem_and_solution}

\end{figure}

To address this limitation, 3DGS and its variants~\cite{kerbl2023splatting,qin2024langsplat, huang2025trigs, chen2025scenesplat, yu2024mip} offer a powerful solution. They achieve real-time, high-quality rendering by using an explicit representation and an efficient GPU rasterization pipeline~\cite{wang2024diffusion}. However, the strong performance of 3DGS usually relies on dense multi-view inputs with sufficient view overlap. In practice, when only a handful of images are available, reconstruction must proceed under sparse-view conditions, often with wide baselines and limited overlap~\cite{charatan2024pixelsplat, zhu2024fsgs, tang2024hisplat, jena2024sparsplat, zheng2024gpsgaussian, fan2024instantsplat, xiong2023sparsegs, wang2024freesplat, xu2025freesplatter, yin2024fewviewgs, shen2025surags, kong2025generative, yang2024gaussianobject}. This scarcity reduces geometric cues and thus impedes the formation of coherent 3D structures. Therefore, generalizing 3DGS to sparse-view regimes has become a key and rapidly evolving research focus~\cite{luo2024review, li2024gsreview}.

Despite the importance of this problem, existing mitigation strategies remain inadequate~\cite{li2024gsreview, younis2025survey}. First, methods that inject geometric priors from depth estimators~\cite{chung2024depth, kumar2024few, Turkulainen2025DNSplatter, huang2025fatesgs} or leverage foundation models for indirect perceptual guidance~\cite{chen2024interngs, zhang2024sags} seldom provide pixel-level, cross-view anchors that directly tie corresponding 3D primitives, which are needed for strong geometric consistency. Second, even techniques that enforce intrinsic consistency~\cite{zhang2024corgs, han2024binocular, park2024dropgaussian} are fundamentally fragile, as their success depends on initial feature matches that are unreliable in sparse settings. Taken together, these shortcomings reflect a common difficulty: sparse-view reconstruction is highly under-constrained, and 2D photometric supervision alone is often insufficient to disambiguate geometry under occlusion and view-dependent appearance.
Moreover, reliable cross-view correspondences are harder to obtain when overlap is limited, which further weakens any consistency regularization that depends on matches~\cite{wu2024cocasplat}. Consequently, as shown in Fig.~\ref{fig:problem_and_solution}, reconstructions exhibit severe artifacts, including distorted 3D shapes and floating debris.

Our work addresses this limitation by complementing 2D rendering losses with correspondence-grounded 3D consistency constraints. Inspired by the observation that sparse-view ambiguities require a supervisory signal that understands what a scene is rather than only what it looks like, we leverage semantic understanding to guide the learning of geometrically coherent 3D structure. Accordingly, we design training to resolve geometric ambiguities by jointly enforcing cross-view consistency over 3D position, shape, and appearance.

Concretely, we present MAC-Splat, a framework grounded in Semantically-Guided Geometric Consistency. The approach integrates a robust correspondence stage with our core contribution, the MAC loss, which leverages semantically augmented correspondence features to enforce holistic 3D consistency. Specifically, we extract sparse 2D correspondences using a strong geometric backbone. We then enrich these matches with part-level features from a frozen DINOv3 encoder via a lightweight Residual Semantic Fusion module. This semantic augmentation improves correspondence reliability and confidence estimation, which in turn strengthens the downstream 3D consistency supervision. The loss then lifts these 2D anchors into 3D and enforces direct, confidence-weighted consistency on the corresponding Gaussians in a shared world frame, using ground-truth poses during training and MASt3R-estimated poses at test time. It jointly regularizes Gaussian centroids for positional consistency, covariance parameters for shape consistency, and opacity and color for appearance consistency. This supervision resolves ambiguities that evade conventional losses and yields reconstructions that are both geometrically coherent and semantically plausible. As shown in Fig.~\ref{fig:problem_and_solution}, matched pixels help maintain consistent Gaussian attributes and reduce common artifacts. In summary, our main contributions are threefold:
\begin{itemize}[leftmargin=*,topsep=1pt, partopsep=0pt, itemsep=2pt]
\item We introduce a novel paradigm, Semantically-Guided Geometric Consistency, which strategically uses semantic information to resolve ambiguities in the 3D optimization process for sparse-view reconstruction.
\item We propose the MAC loss, a comprehensive objective that is empowered by semantically-augmented correspondences to jointly regularize the 3D position, shape, and appearance of Gaussian primitives.
\item We design a Residual Semantic Fusion module that enriches geometric features with semantic priors, providing the foundational representation that enables our semantically-aware consistency objective.
\end{itemize}

\section{Related Work}
\label{sec:related_work}
NeRF~\cite{mildenhall2020nerf} and 3DGS~\cite{kerbl2023splatting} have demonstrated remarkable performance in 3D reconstruction and novel view synthesis. However, both paradigms traditionally rely on densely captured, accurately posed images and per-scene optimization, which limits scalability and real-world deployment. In practical scenarios such as mobile AR/VR, robotics, or clinical imaging~\cite{Matsuki2024GSSLAM, Keetha2024SplaTAM, Yan2024GSSLAM}, only a few views may be available, and camera poses can be unreliable. This challenge has driven the transition from per-scene optimization to generalizable feed-forward models capable of predicting 3D representations directly from sparse inputs with minimal test-time cost~\cite{chung2024depthreg, kumar2024few, zhu2024fsgs, younis2025survey}.

\noindent \textbf{\textit{Generalizable Feed-Forward 3DGS.}}
Recent advances replace per-scene optimization with feed-forward networks that predict 3D Gaussians in a single pass. Early methods such as PixelSplat~\cite{charatan2024pixelsplat} and MVSplat~\cite{chen2024mvsplat} assume accurate SfM poses, while pose-free approaches (e.g., Splatt3R~\cite{lentsch2024splatt3r}, NoPoSplat~\cite{sella2024noposplat}, AnySplat~\cite{jiang2025anysplat}, SPFSplat~\cite{huang2025no}) extend the paradigm to uncalibrated inputs by leveraging foundation models such as DUSt3R~\cite{wang2024dust3r} or MASt3R~\cite{lentsch2024mast3r} to predict a canonical-space representation and infer or bypass poses. Despite improved practicality, sparse-view and wide-baseline reconstruction remains ill-posed: 2D rendering supervision is often insufficient to resolve occlusion and view-dependent ambiguity, and correspondence quality degrades with limited overlap, allowing matching errors to propagate into 3D Gaussians and produce floaters or duplicated surfaces. This motivates explicit 3D regularization beyond feed-forward prediction.

\noindent \textbf{\textit{External Geometric Priors.}}
Many such priors are applied either in posed pipelines or require pose estimates to align multi-view cues. A prominent line of work stabilizes sparse-view optimization by injecting explicit geometric cues. Depth and normal maps predicted by monocular estimators are frequently used to regularize Gaussian placement and suppress floaters~\cite{chung2024depth, kumar2024few, ming2021deep, Turkulainen2025DNSplatter}. Some approaches further integrate multi-view stereo signals or surface extraction pipelines to improve structural coherence~\cite{chen2024mvsplat, liu2024mvsgaussian, Guedon2024SuGaR}. These priors effectively constrain macroscopic surface layout and improve reconstruction stability. However, they provide proxy geometry rather than explicitly binding corresponding 3D primitives across views, and their effectiveness depends on the reliability of external estimators and pose alignment. Monocular predictions may introduce scale ambiguity or bias in textureless or specular regions, while multi-view stereo cues degrade under sparse overlap or wide baselines.

\noindent \textbf{\textit{Representation-Level and Foundation Priors.}}
Another direction leverages large-scale foundation models to enhance reconstruction robustness. One line of work introduces diffusion-based priors to regularize rendered views toward realistic image statistics~\cite{wang2024diffusion, meng2024auggs, liu20243dgs}. Another line uses vision transformer features or vision-language signals to provide richer semantic descriptors for reconstruction networks~\cite{li2024gslrm, zou2024triplane, chen2024interngs, zhang2024sags, li2024geogaussian, yu2024lm, chen2025slgaussian, Liu_2025_ICCV}. These priors often improve global appearance coherence and feature distinctiveness, especially in visually ambiguous regions. However, they typically operate at the representation or image-projection level and do not explicitly couple corresponding 3D primitives across views. As a result, identity-level and attribute-level consistency is addressed only indirectly through image-space supervision.

\noindent \textbf{\textit{Intrinsic Cross-View Regularization.}}
Beyond injecting external cues, several works introduce intrinsic training objectives to promote structural robustness. Stochastic Gaussian dropout reduces overfitting~\cite{park2024dropgaussian, peng2024structure, zhao2025self}, and binocular guidance or multi-view surface constraints encourage view-consistent predictions~\cite{han2024binocular, li2025mvgsr}. Although effective, these objectives typically enforce agreement at the projection or surface level. Explicit identity-level coupling between matched 3D primitives is less commonly enforced, so primitive attributes are still not directly constrained to be consistent across views.

The preceding analysis reveals a common gap in sparse-view generalizable 3DGS: most methods improve reconstruction via proxy geometry, image-level priors, or projection-level consistency, but rarely enforce explicit identity-level coupling between corresponding 3D primitives across views. MAC-Splat addresses this gap by first improving correspondence reliability using a strong geometric backbone augmented with DINOv3 features, and then applying a Multi-Attribute Consistency loss that directly regularizes matched Gaussians in 3D, jointly aligning their position, shape, and appearance.

\section{Methodology}
\label{sec:method}

\begin{figure*}[t]
    \centering
    \includegraphics[width=\textwidth]{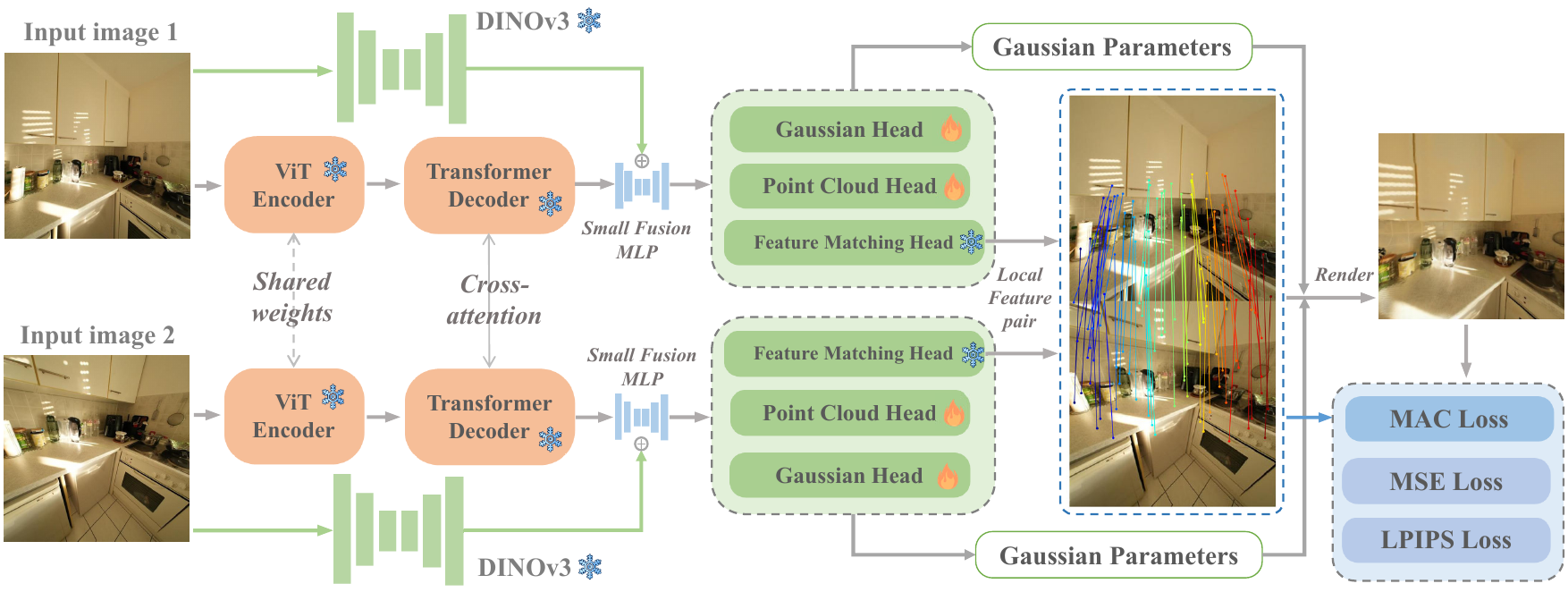}
    \caption{\textbf{Architectural overview of MAC-Splat.}
    The framework augments a MASt3R backbone with features from a DINOv3 encoder through a residual fusion module to produce semantically enriched descriptors. A multi-stage filtering procedure yields sparse, high-confidence 2D correspondences. These matches anchor the Multi-Attribute Consistency (MAC) loss, which compares the world-space position, shape, and appearance of the associated Gaussians.}
    \label{fig:framework}

\end{figure*}

Our methodology targets sparse-view 3D reconstruction, where limited viewpoints introduce geometric ambiguity. We decompose the task into coupled components: establishing high-fidelity 2D correspondences and leveraging these correspondences to enforce direct 3D geometric consistency. The proposed MAC-Splat pipeline (Fig.~\ref{fig:framework}) integrates a feature fusion module, a correspondence filtering procedure, and a multi-attribute 3D consistency loss to realize this.

\subsection{High-Fidelity Correspondence Establishment}
\label{sec:correspondence}
To supply reliable geometric evidence for 3D consistency, we first construct a compact yet highly trustworthy set of 2D anchors that capture strong cross-view relations. Because accurate 3D supervision hinges on these anchors, we design a multi-stage procedure that progressively filters and refines candidate matches, yielding a sparse set of high-confidence correspondences $\mathcal{M}$ for each image pair. Concretely, we use MASt3R~\cite{lentsch2024mast3r} as the geometric matching backbone and enhance its descriptors with a lightweight residual semantic fusion module. This module injects dense semantic features from a frozen DINOv3~\cite{simeoni2024dinov3} encoder, providing corrective cues in ambiguous regions. Given a baseline descriptor $f_{\text{dec}}$ from the MASt3R decoder and a spatially aligned DINOv3 feature $f_{\text{dino}}$, a small fusion MLP $g_{\text{fuse}}$ predicts a residual
$r = g_{\text{fuse}}([f_{\text{dec}}, f_{\text{dino}}])$,
which is added to the baseline descriptor,
$f' = f_{\text{dec}} + r$. DINOv3 features are resized to the decoder grid via bilinear interpolation and projected to the descriptor dimension with a linear layer before fusion. The resulting descriptor maps for the two views,
$\mathcal{F}_1$ and $\mathcal{F}_2$, are more distinctive while remaining compatible with the original MASt3R features. Residual fusion is applied only at the final decoder layer, using a single MLP shared across spatial locations. We freeze the MASt3R matching head weights to preserve pretrained behavior, but reciprocal NN matches depend on the input descriptors, so feeding fused descriptors $f'$ changes the correspondences.

Building on these enhanced descriptors, we generate dense candidate correspondences by identifying reciprocal nearest neighbors between $\mathcal{F}_1$ and $\mathcal{F}_2$.~A pair of pixels $(\boldsymbol{p}_i^1, \boldsymbol{p}_j^2)$ from images $\mathbf{I}_1$ and $\mathbf{I}_2$ constitutes a reciprocal match if their respective descriptors $f'_{1,i} \in \mathcal{F}_1$ and $f'_{2,j} \in \mathcal{F}_2$
satisfy,
\begin{equation}
i = \arg\min_{k} D(f'_{2,j}, f'_{1,k}), \quad
j = \arg\min_{k} D(f'_{1,i}, f'_{2,k}),
\end{equation}
where $D(\cdot,\cdot)$ denotes cosine distance in feature space.
This bidirectional check enforces a symmetry constraint and effectively prunes many ambiguous matches, especially in occluded or repetitive regions.

Finally, we refine the correspondences using the network’s predicted descriptor confidence. Following MASt3R~\cite{lentsch2024mast3r}, we use a confidence head shared across all pixels, which is frozen and used only for correspondence gating. For each descriptor $f'$, this head outputs a scalar confidence value $c(f')$ that estimates its distinctiveness. We define the joint confidence of a match as the product of the confidences at its two endpoints,
$w_{ij} = c(f'_{1,i})\, c(f'_{2,j})$,
and we retain only those matches that satisfy $w_{ij} > \tau_{\text{conf}}$, thereby discarding correspondences in regions where the network expresses low certainty.
Because the retained anchors are softly weighted by their joint confidence in the subsequent MAC loss and residual outliers are suppressed by a Huber penalty, our framework is highly robust to the exact choice of $\tau_{\text{conf}}$. Furthermore, if no match survives this gating in extremely ambiguous regions, the MAC loss gracefully falls back to zero, allowing the network to rely solely on photometric supervision.
As a result, the final output is a sparse set of high-fidelity 2D matches,
$\mathcal{M} = \{\boldsymbol{p}_k^1 \leftrightarrow \boldsymbol{p}_k^2\}_{k=1}^K$,
which serve as geometric anchors for the subsequent MAC loss on 3D Gaussian attributes. We construct $\mathcal{M}$ for each sampled image pair; in practice, using multiple pairs per scene provides overlapping anchors for the same Gaussians.

\subsection{Direct Multi-Attribute 3D Consistency Loss}
\label{sec:mac_loss}
\begin{figure*}[t]
    \centering
    \includegraphics[width=\textwidth]{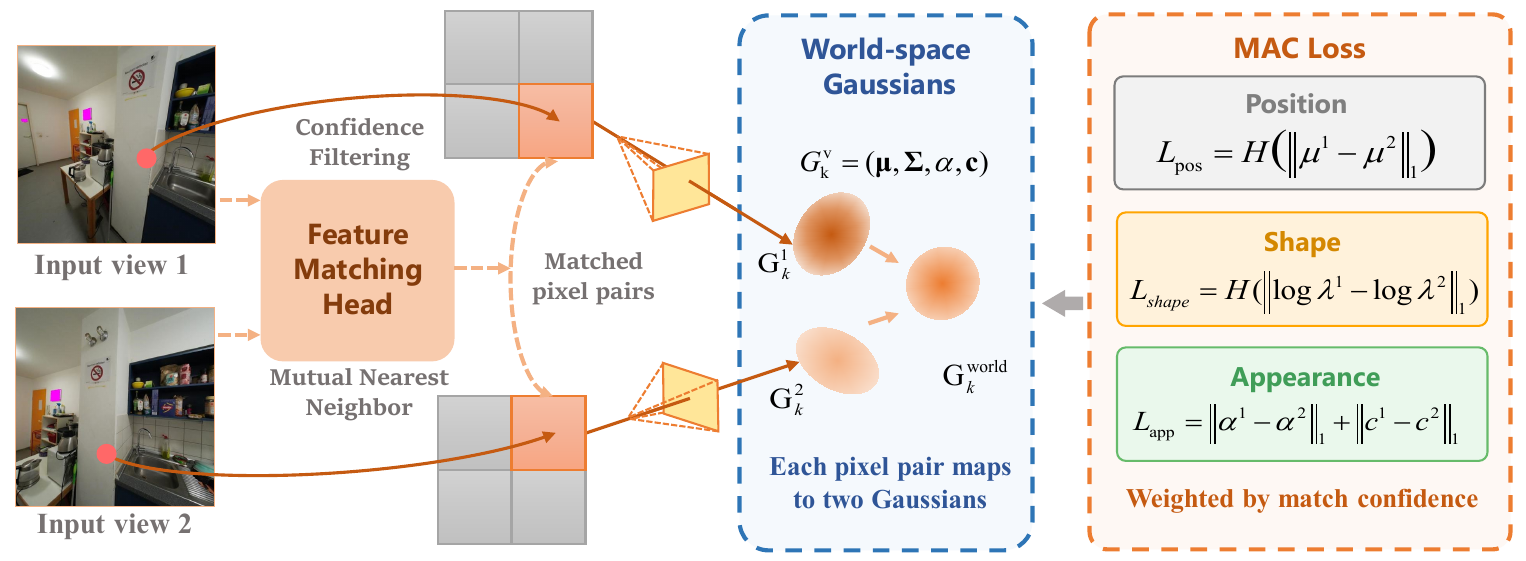}
    \caption{\textbf{From matched pixels to MAC loss.}
    Each matched pixel pair selects two Gaussians in world space, and the MAC loss enforces consistency in their position, shape, and appearance.}
    \label{fig:mac_loss}

\end{figure*}

Given the 2D correspondences $\mathcal{M}$ from Section~\ref{sec:correspondence}, we define a Multi-Attribute Consistency loss on 3D Gaussians.
The loss uses the matches in $\mathcal{M}$ as anchors and penalizes differences in position, shape, and appearance between the Gaussians associated with each matched pixel pair. During training, we use ground-truth camera-to-world transformations $T_{wc}^v$ to compare all attributes in a common world coordinate system; at test time, $T_{wc}^v$ is obtained from MASt3R pose estimation. An overview of this pipeline, from matched pixels to world-space Gaussians and the corresponding MAC loss terms, is provided in Fig.~\ref{fig:mac_loss}.

\noindent\textbf{From matched pixels to Gaussian pairs.}
For each training pair $(\mathbf{I}_1,\mathbf{I}_2)$, our Gaussian prediction head predicts, for each view, a dense map of Gaussian parameters aligned with the descriptor grid.
At every location $(u,v)$ in view $v$, we obtain a 3D centroid $\boldsymbol{\mu}_{\text{cam}}^v(u,v)\in\mathbb{R}^3$, a covariance matrix $\mathbf{\Sigma}_{\text{cam}}^v(u,v)\in\mathrm{Sym}^+(3)$, an opacity $\alpha^v(u,v)$, and spherical harmonic coefficients $\boldsymbol{c}^v(u,v)$. The reciprocal matches in $\mathcal{M}$ are defined on the same grid: a match $(\boldsymbol{p}_k^1,\boldsymbol{p}_k^2)$ specifies two pixel coordinates in the parameter maps of $I_1$ and $I_2$, together with a batch index. For these coordinates, we obtain the associated Gaussians $G_k^1$ and $G_k^2$ by bilinearly sampling the predicted parameter maps using differentiable grid sampling. This yields, for each correspondence $k$, a pair of Gaussians,
\[
G_k^v = \bigl(\boldsymbol{\mu}_{\text{cam}}^{v,k},\,\mathbf{\Sigma}_{\text{cam}}^{v,k},\,\alpha^{v,k},\,\boldsymbol{c}^{v,k}\bigr), \quad v\in\{1,2\},
\]
where the superscript $k$ denotes sampling at $\boldsymbol{p}_k^v$. In the following, we transform these parameters into the world coordinate system using the camera-to-world transforms $T_{wc}^1$ and $T_{wc}^2$, and then apply the positional, shape, and appearance consistency terms.

\noindent(\textit{Positional consistency.})
We encourage the 3D centroids of matched Gaussians to coincide in world space. Given the camera-to-world transform
$T_{wc}^v=[\mathbf{R}^v\ \boldsymbol{t}^v]$, the world-space centroid of $G_k^v$ is,
\begin{equation}
    \boldsymbol{\mu}_{\text{world}}^{v,k} = \mathbf{R}^v \boldsymbol{\mu}_{\text{cam}}^{v,k} + \boldsymbol{t}^v.
\end{equation}
For each correspondence, we define a robust positional loss,
\begin{equation}
    \loss{pos}^{(k)} = \mathcal{H}\bigl(\|\boldsymbol{\mu}_{\text{world}}^{1,k} - \boldsymbol{\mu}_{\text{world}}^{2,k}\|_1\bigr),
\end{equation}
which is evaluated only when both centroids lie in front of their cameras ($z_{\text{cam}}\!\!>\!\!0$). Here, $\mathcal{H}$ denotes the Huber loss.

\noindent(\textit{Shape Consistency.})
To promote consistent 3D shape across views, we compare the covariance matrices that define each Gaussian. Since each covariance $\mathbf{\Sigma}\in\mathrm{Sym}^+(3)$ is predicted in the camera frame, we first transfer it to the world frame using the rotation $\mathbf{R}$ from $T_{wc}$,
\begin{equation}
    \mathbf{\Sigma}_{\text{world}} = \mathbf{R}\,\mathbf{\Sigma}_{\text{cam}}\,\mathbf{R}^\top.
\end{equation}
A Euclidean distance between covariance matrices does not respect the geometry of $\mathrm{Sym}^{\!+}(3)$ and is not invariant to rotations or uniform scaling. To address this, we compare the logarithms of their eigenvalues.
For a matched pair $\!(G_k^1,\!G_k^2)\!$ with eigenvalues $\boldsymbol{\lambda}^{\!1,k}$ and $\boldsymbol{\lambda}^{\!2,k}\!\!$, the shape loss is,
\begin{equation}
    \begin{aligned}
        \loss{shape}^{(k)} &= \mathcal{H}\!\left(\left\|\log\boldsymbol{\lambda}^{1,k}-\log\boldsymbol{\lambda}^{2,k}\right\|_1\right),\\
        \boldsymbol{\lambda}^{v,k} &= \mathrm{eig}\!\left(\mathbf{\Sigma}_{\text{world}}^{v,k}\right),\quad v\in\{1,2\}.
    \end{aligned}
\end{equation}
This metric is rotation invariant because it depends only on the eigenvalues, and it measures relative differences in the principal axes through $\log(a/b)=\log(a)-\log(b)$. Consequently, this formulation is invariant to uniform global scaling and focuses exclusively on mismatches in anisotropy. By penalizing relative scale ratios rather than absolute scale differences, it actively prevents spurious stretching or flattening of the geometry across views.

\noindent(\textit{Appearance Consistency.})
We regularize the photometric attributes by penalizing differences in opacity and spherical harmonic coefficients,
\begin{equation}
    \loss{app}^{(k)} = \|\alpha^{1,k}-\alpha^{2,k}\|_1 + \|\boldsymbol{c}^{1,k}-\boldsymbol{c}^{2,k}\|_1.
\end{equation}

\noindent\textbf{Multi-Attribute Consistency Loss.}
The MAC loss combines the positional, shape, and appearance terms using the joint confidence $w_k$ of each correspondence,
\begin{equation}
    \loss{MAC} = 
    \frac{\sum_{k=1}^{K} w_k\bigl(\lambda_p \loss{pos}^{(k)} + \lambda_s \loss{shape}^{(k)} + \lambda_a \loss{app}^{(k)}\bigr)}
         {\sum_{k=1}^{K} w_k + \epsilon}.
\end{equation}
Here, $w_k$ denotes the joint confidence of the $k$-th matched pixel pair, $\lambda_p,\lambda_s,\lambda_a$ are scalar weights for the three terms, and $\epsilon$ is a small constant that prevents division by zero.

\noindent\textbf{Preventing Geometric Degeneracy.}
A critical theoretical property of our log-eigenvalue shape loss is its gradient behavior near geometric degeneracy. For a Huber-penalized log-difference, the magnitude of the gradient with respect to a single eigenvalue $\lambda$ scales inversely with the eigenvalue itself:
\begin{equation}
    \left\lvert \frac{\partial \loss{shape}}{\partial \lambda} \right\rvert \propto \frac{1}{\lambda}.
\end{equation}
As a Gaussian primitive becomes degenerate—meaning its smallest eigenvalue $\lambda_{\min}$ approaches zero and it collapses into a flat or needle-like shape—the gradient magnitude sharply increases. This creates a strong repulsive effect that actively pushes $\lambda_{\min}$ away from zero. This intrinsic anti-degeneracy mechanism heavily stabilizes the optimization landscape, explicitly discouraging the collapsed floating artifacts that frequently plague sparse-view reconstructions guided solely by photometric gradients.

\subsection{Overall Training Objective}
\label{sec:total_loss}

Having defined the MAC loss, we now describe the overall training objective of our model. We train the Gaussian prediction head and the residual fusion module with a combination of photometric loss and the proposed MAC loss, while keeping DINOv3 and the MASt3R matching heads frozen,
\begin{equation}
    \loss{total} = \loss{photo} + \lambda_{\text{MAC}} \loss{MAC}.
\end{equation}
Here, $\loss{photo}$ is a weighted sum of an $\ell_2$ reconstruction term and an LPIPS term~\cite{zhang2018unreasonable}, both applied to target views rendered from the predicted Gaussians. The scalar $\lambda_{\text{MAC}}$ controls the strength of the 3D consistency regularization and is kept fixed across all experiments. Although the MASt3R correspondence loss can be added as an auxiliary term in principle, we set its weight to zero in all reported results.

\section{Experiments}
\label{sec:experiments}

We evaluate MAC-Splat along four aspects: (i) quantitative and perceptual performance compared to state-of-the-art generalizable methods; (ii) the contribution of the MAC loss and residual semantic fusion; (iii) robustness under varying viewpoint separation; and (iv) zero-shot generalization from ScanNet++ to DTU.

\subsection{Experimental Setup}
\label{sec:setup}

\noindent\textbf{Dataset and Difficulty Splits.}
We use ScanNet++~\cite{yeshwanth2023scannet++}, a large-scale indoor dataset with accurate ground-truth geometry. Following Splatt3R~\cite{lentsch2024splatt3r}, we group test pairs into four difficulty subsets using coverage thresholds $(\phi,\psi)$: \textbf{Close}, \textbf{Medium}, \textbf{Wide}, and \textbf{Very Wide}. These thresholds measure the fraction of depth-consistent visible pixels between context and target views. Larger $(\phi,\psi)$ values correspond to shorter baselines and higher view overlap, while lower values indicate significantly wider baselines and minimal overlap, providing a rigorous testbed for geometric robustness.

\begin{table*}[t!]

    \centering
    \caption{\textbf{Quantitative results on ScanNet++.} We report PSNR ($\uparrow$), SSIM ($\uparrow$), and LPIPS ($\downarrow$) across four difficulty splits controlled by thresholds $(\phi,\psi)$ (higher means more overlap and smaller baseline). Best results are in \textbf{bold}.}
    \label{tab:main_results}
    \resizebox{\textwidth}{!}{%
    \begin{tabular}{l|ccc|ccc|ccc|ccc}
        \toprule
        \multirow{2}{*}{\textbf{Method}} & \multicolumn{3}{c|}{\textbf{Close}} & \multicolumn{3}{c|}{\textbf{Medium}} & \multicolumn{3}{c|}{\textbf{Wide}} & \multicolumn{3}{c}{\textbf{Very Wide}} \\
        & \multicolumn{3}{c|}{$(\phi{=}0.9,\ \psi{=}0.9)$} & \multicolumn{3}{c|}{$(\phi{=}0.7,\ \psi{=}0.7)$} & \multicolumn{3}{c|}{$(\phi{=}0.5,\ \psi{=}0.5)$} & \multicolumn{3}{c}{$(\phi{=}0.3,\ \psi{=}0.3)$} \\
        \cmidrule(lr){2-4} \cmidrule(lr){5-7} \cmidrule(lr){8-10} \cmidrule(lr){11-13}
        & PSNR & SSIM & LPIPS & PSNR & SSIM & LPIPS & PSNR & SSIM & LPIPS & PSNR & SSIM & LPIPS \\
        \midrule
        Splatt3R~\cite{lentsch2024splatt3r}              & 18.89 & 0.757 & 0.233 & 18.79 & 0.760 & 0.228 & 16.71 & 0.686 & 0.289 & 16.16 & 0.662 & 0.289 \\
        MASt3R~\cite{lentsch2024mast3r}                  & 18.44 & 0.722 & 0.292 & 15.23 & 0.648 & 0.359 & 15.72 & 0.719 & 0.295 & 15.82 & 0.608 & 0.278 \\
        DUSt3R~\cite{wang2024dust3r}                     & 16.46 & 0.693 & 0.279 & 15.96 & 0.652 & 0.312 & 15.64 & 0.704 & 0.323 & 15.23 & 0.612 & 0.262 \\
        NoPoSplat~\cite{sella2024noposplat}              & 19.97 & 0.722 & 0.251 & 19.66 & 0.801 & 0.263 & 18.62 & 0.745 & 0.276 & 18.74 & 0.730 & 0.239 \\
        MVSplat~\cite{chen2024mvsplat}                   & 22.64 & 0.802 & 0.192 & 18.32 & 0.757 & 0.287 & 16.11 & 0.710 & 0.295 & 13.91 & 0.662 & 0.371 \\
        PixelSplat~\cite{charatan2024pixelsplat}         & \textbf{23.98} & \textbf{0.817} & 0.169 & 20.31 & 0.783 & 0.227 & 18.46 & 0.770 & 0.235 & 15.36 & 0.673 & 0.329 \\ 
        \midrule
        \textbf{MAC-Splat (Ours)}                        & 22.76 & 0.810 & \textbf{0.146} & \textbf{22.60} & \textbf{0.816} & \textbf{0.135} & \textbf{22.09} & \textbf{0.818} & \textbf{0.142} & \textbf{21.34} & \textbf{0.818} & \textbf{0.158} \\
        \bottomrule
    \end{tabular}
    }
     
\end{table*}

\begin{figure*}[t!]
    \centering
    \includegraphics[width=\textwidth]{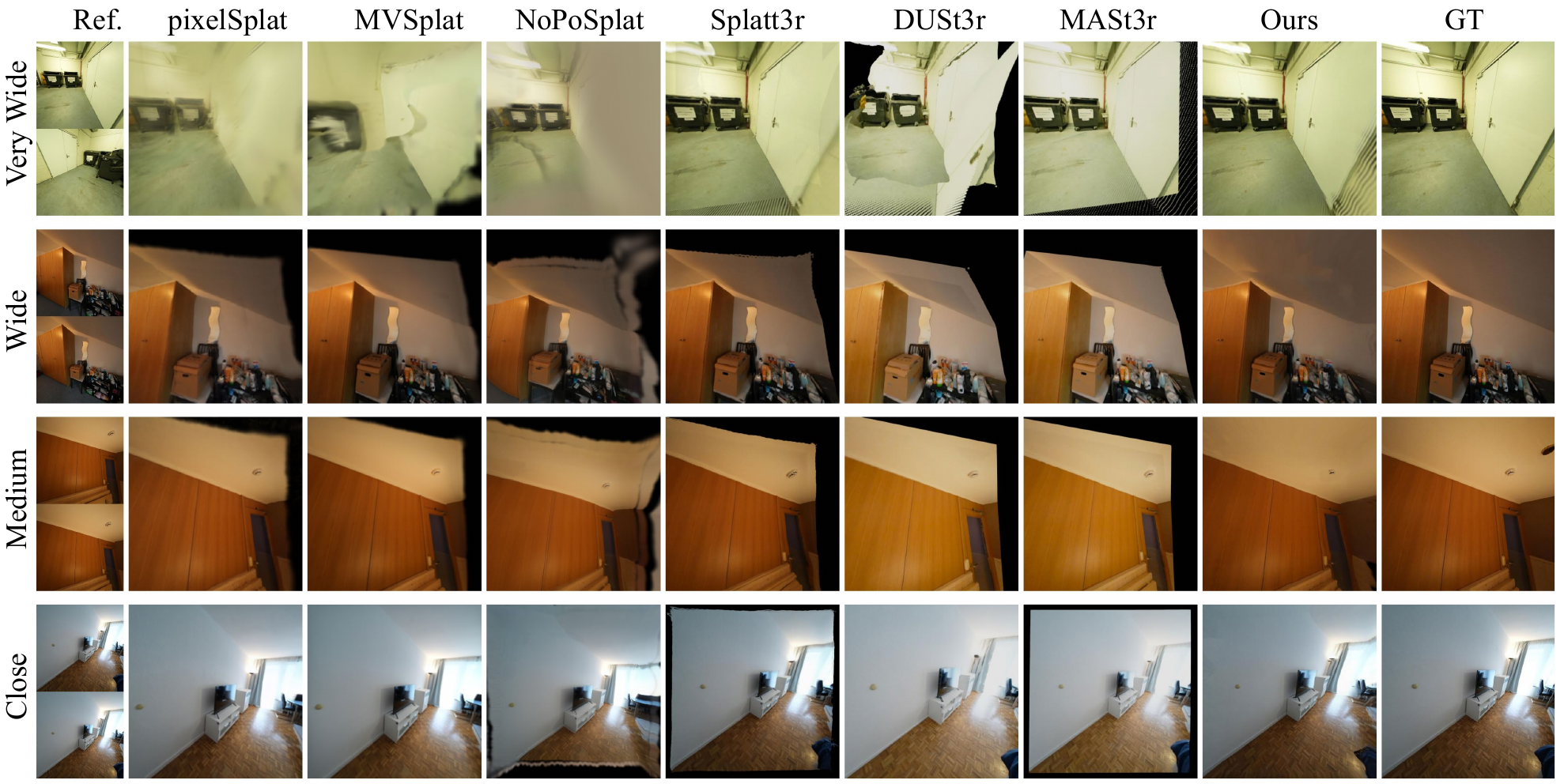}
    \caption{\textbf{Qualitative comparison on ScanNet++.} Compared to recent methods, MAC-Splat significantly reduces severe floaters and ghosting artifacts. By enforcing 3D multi-attribute consistency on high-fidelity anchors, it produces sharper textures and cleaner occlusion boundaries under challenging wide-baseline conditions.}
    \label{fig:qualitative_main}

\end{figure*}

\noindent\textbf{Inference protocol.}
For MAC-Splat, during inference the camera poses for the input context views are estimated by the MASt3R backbone from the input views, without ground-truth poses or external SfM for the input; all baselines are evaluated under their authors' official inference protocols, including their pose requirements.
Ground-truth poses are used only to define target novel-view cameras for metric computation, which is standard for evaluating novel view synthesis.

\noindent\textbf{Cross-dataset protocol.}
For DTU~\cite{aanas2016dtu} evaluation, MAC-Splat operates in a strictly zero-shot setting, using the model trained exclusively on ScanNet++ without any dataset-specific fine-tuning. All baseline methods are similarly evaluated using their authors' official pretrained models. To control evaluation variables, all methods use the official DTU camera intrinsics and a common input resolution, and we report metrics averaged over all target views.

\noindent\textbf{Evaluation metrics.}
We report peak signal-to-noise ratio (PSNR), structural similarity (SSIM), and LPIPS~\cite{zhang2018unreasonable}, averaged over all target views and test scenes. All metrics are computed in sRGB space on the full image unless otherwise specified. Furthermore, following Splatt3R~\cite{lentsch2024splatt3r}, we also report \textbf{masked metrics} evaluated exclusively on observable regions.

\noindent\textbf{Comparison methods.}~We compare MAC-Splat with correspondence-based baselines DUSt3R~\cite{wang2024dust3r} and MASt3R~\cite{lentsch2024mast3r}, which are rendered as colored point clouds. We also evaluate generalizable 3D Gaussian splatting methods, including Splatt3R~\cite{lentsch2024splatt3r}, MVSplat~\cite{chen2024mvsplat}, PixelSplat~\cite{charatan2024pixelsplat}, and NoPoSplat~\cite{sella2024noposplat}, using their official implementations under a common evaluation protocol.

\noindent\textbf{Implementation details.}
MAC-Splat is trained on ScanNet++ using four NVIDIA RTX~4090 GPUs and AdamW with a learning rate of $10^{-5}$.
The overall MAC loss weight is set to $\lambda_{\text{MAC}} = 0.25$, with internal weights $\lambda_p = 1.0$, $\lambda_s = 0.02$, and $\lambda_a = 0.1$ for the positional, shape, and appearance terms.
PixelSplat, MVSplat, and NoPoSplat are evaluated using the publicly released checkpoints provided by the authors.
Splatt3R, DUSt3R and MASt3R are used with their official pretrained weights.
For all baseline methods, we adopt the authors' recommended inference settings without modification.

\begin{table}[t]
\centering
\caption{Cross-dataset results on DTU. All methods are evaluated zero-shot on DTU. We report PSNR ($\uparrow$), SSIM ($\uparrow$), and LPIPS ($\downarrow$), averaged over all target views. Best results are in \textbf{bold}.}
\label{tab:dtu}
\begin{tabular}{lccc}
\toprule
Method & PSNR $\uparrow$ & SSIM $\uparrow$ & LPIPS $\downarrow$ \\
\midrule
PixelSplat~\cite{charatan2024pixelsplat} & 13.33 & 0.517 & 0.329 \\
MVSplat~\cite{chen2024mvsplat}    & 14.03 & 0.522 & 0.334 \\
Splatt3R~\cite{lentsch2024splatt3r}   & 12.13 & 0.454 & 0.437 \\
\textbf{MAC-Splat (Ours)} & \textbf{17.32} & \textbf{0.605} & \textbf{0.311} \\
\bottomrule
\end{tabular}
 
\end{table}

\begin{figure*}[t]
    \centering
    \includegraphics[width=1\textwidth]{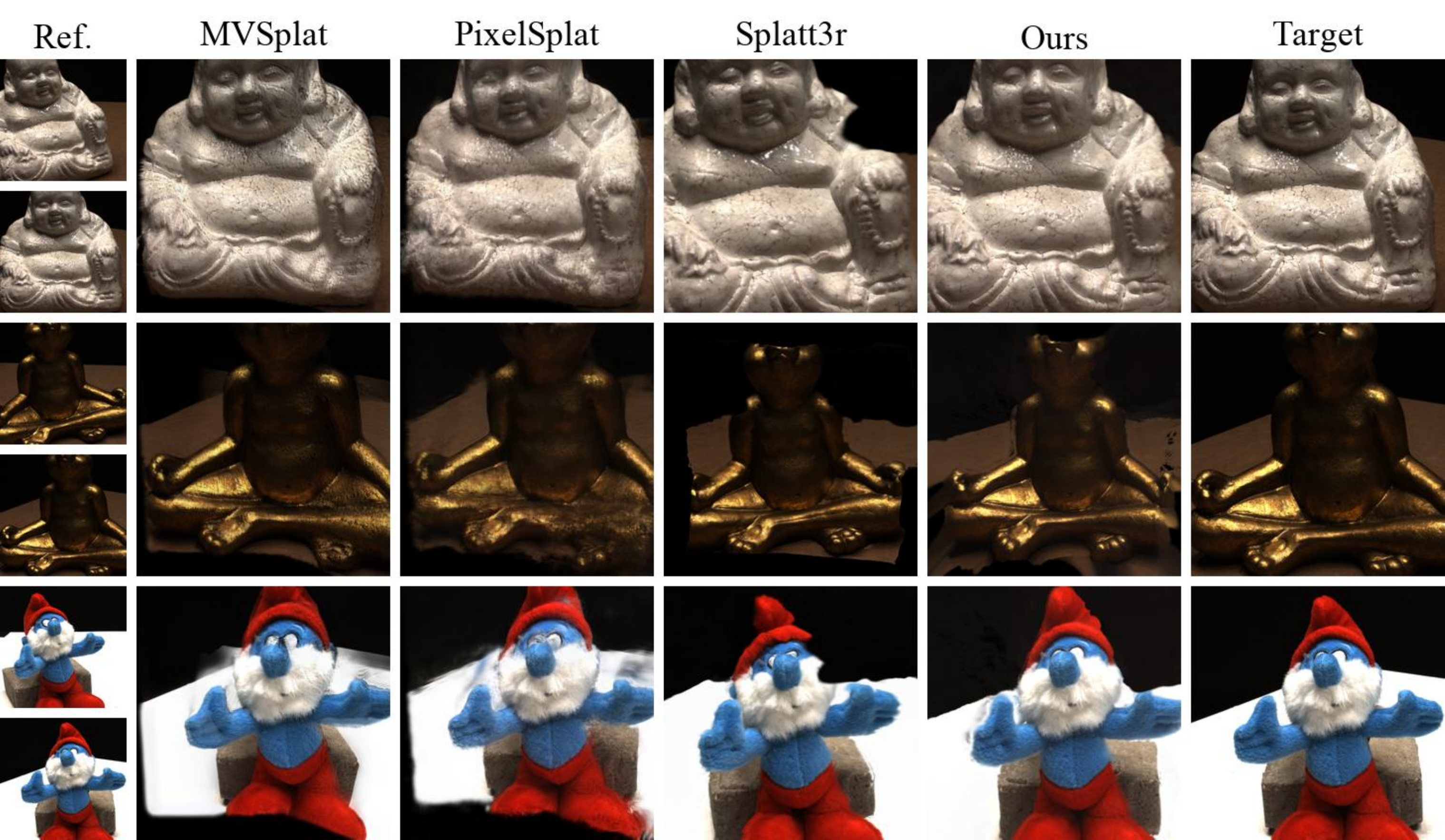}
    \caption{\textbf{Cross-dataset generalization: ScanNet++$\rightarrow$DTU.} Compared to other generalizable 3DGS methods, MAC-Splat better preserves geometry and edges and produces fewer artifacts under the controlled lighting conditions of DTU.}
    \label{fig:cross_dtu}
     
\end{figure*}

\subsection{Quantitative Comparison}
\label{sec:quantitative}

Table~\ref{tab:main_results} reports results on ScanNet++.
Fig.~\ref{fig:qualitative_main} provides qualitative comparisons on representative ScanNet++ scenes.
Under our colored point-cloud rendering protocol, DUSt3R and MASt3R obtain lower image metrics than 3D Gaussian splatting methods.
Among 3DGS methods, PixelSplat achieves the highest PSNR on \textbf{Close} (23.98\,dB) but drops to 15.36\,dB on \textbf{Very Wide}.
In contrast, MAC-Splat decreases from 22.76\,dB to 21.34\,dB and achieves the best PSNR, SSIM, and LPIPS on \textbf{Medium}, \textbf{Wide}, and \textbf{Very Wide}, while attaining the best LPIPS on \textbf{Close}.
NoPoSplat remains competitive across subsets but still lags behind MAC-Splat on \textbf{Wide} and \textbf{Very Wide}, indicating that wide-baseline generalization remains challenging under limited overlap.
Compared to Splatt3R, MAC-Splat improves the average PSNR across the four subsets by more than 4.5\,dB and reduces the average LPIPS by about 44\%.

\noindent\textbf{Robustness to wide baselines.}
A key trend in Table~\ref{tab:main_results} is the behavior across subsets with decreasing overlap (lower $(\phi,\psi)$).
Across the four subsets from \textbf{Close} to \textbf{Very Wide} defined by $(\phi,\psi)$, PixelSplat and MVSplat lose more than 8\,dB in PSNR, while MAC-Splat drops by only 1.42\,dB and maintains the best scores in all three metrics on the harder splits.
LPIPS follows a similar pattern: for PixelSplat and MVSplat it roughly doubles on \textbf{Very Wide}, whereas for MAC-Splat it changes only from 0.146 to 0.158, indicating that perceptual quality is largely preserved under long baselines.
We attribute this robustness to the MAC-Splat design, which uses high-confidence correspondences to couple Gaussians across views and enforces consistency in world space on position, shape, and appearance, rather than relying solely on multi-view photometric supervision.

\begin{table*}[t!]
    \centering
    \caption{\textbf{Quantitative comparison on masked metrics (observable regions).} Metrics are calculated strictly within the Splatt3R visibility mask to evaluate the reconstruction fidelity of observed geometry.}
    \label{tab:masked_results}
    \resizebox{\textwidth}{!}{%
    \begin{tabular}{l|cc|cc|cc|cc}
        \toprule
        \multirow{2}{*}{\textbf{Method}} & \multicolumn{2}{c|}{\textbf{Close}} & \multicolumn{2}{c|}{\textbf{Medium}} & \multicolumn{2}{c|}{\textbf{Wide}} & \multicolumn{2}{c}{\textbf{Very Wide}} \\
        & \multicolumn{2}{c|}{$(\phi{=}0.9,\ \psi{=}0.9)$} & \multicolumn{2}{c|}{$(\phi{=}0.7,\ \psi{=}0.7)$} & \multicolumn{2}{c|}{$(\phi{=}0.5,\ \psi{=}0.5)$} & \multicolumn{2}{c}{$(\phi{=}0.3,\ \psi{=}0.3)$} \\
        \cmidrule(lr){2-3} \cmidrule(lr){4-5} \cmidrule(lr){6-7} \cmidrule(lr){8-9}
        & PSNR & LPIPS & PSNR & LPIPS & PSNR & LPIPS & PSNR & LPIPS \\
        \midrule
        Splatt3R~\cite{lentsch2024splatt3r}              & 14.73 & 0.233 & 14.44 & 0.242 & 13.79 & 0.251 & 13.04 & 0.258 \\
        MASt3R~\cite{lentsch2024mast3r}                  & 13.57 & 0.283 & 12.96 & 0.280 & 12.50 & 0.293 & 11.27 & 0.322 \\
        \midrule
        \textbf{MAC-Splat (Ours)}                        & \textbf{21.56} & \textbf{0.159} & \textbf{21.06} & \textbf{0.163} & \textbf{20.26} & \textbf{0.178} & \textbf{19.06} & \textbf{0.189} \\
        \bottomrule
    \end{tabular}
    }
     
\end{table*}

\noindent\textbf{Evaluation on Observable Regions.}
To further verify that the gains reflect improved reconstruction of geometry that is actually constrained by the inputs, we additionally evaluate using the visibility-mask definition proposed by Splatt3R~\cite{lentsch2024splatt3r}. This mask restricts evaluation to pixels that are visible from at least one context view and satisfy depth-consistency, thereby reducing the influence of unobserved regions where image-space completion can dominate the full-image metrics. As shown in Table~\ref{tab:masked_results}, MAC-Splat outperforms Splatt3R and MASt3R on masked PSNR/LPIPS across all difficulty splits. Notably, on the \textbf{Very Wide} subset, Splatt3R drops to 13.04\,dB masked PSNR, whereas MAC-Splat reaches 19.06\,dB, yielding a +6.02\,dB gap in regions that are jointly observable. This improvement is consistent with our design goal: the MAC loss explicitly couples matched Gaussians across views and enforces multi-attribute consistency in 3D, which reduces geometric drift and suppresses double surfaces even when overlap is minimal.

\begin{figure*}[t!]
    \centering
    \includegraphics[width=\textwidth]{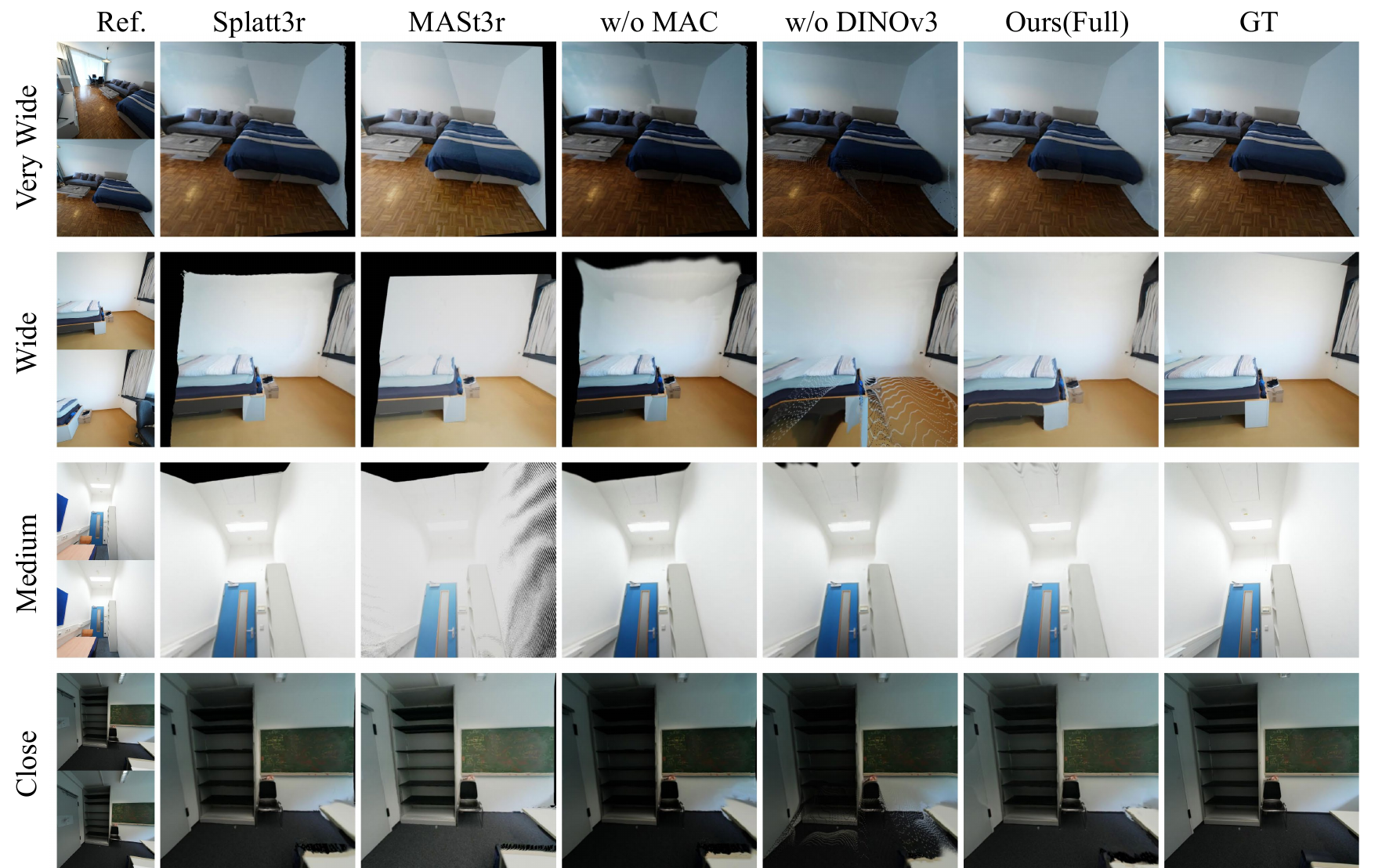}
    \caption{\textbf{Qualitative ablation on ScanNet++.} Pure 2D photometric supervision (\textit{w/o MAC Loss}) yields severe geometric distortions and floaters. While the MAC loss alone (\textit{w/o DINOv3}) corrects global 3D geometry, our full model uses semantically enriched correspondences to further refine local details and cleanly preserve thin structures.}
    \label{fig:qualitative_ablation}

\end{figure*}

\subsection{Cross-Dataset Generalization}
\label{sec:cross_dtu}
\noindent\textbf{Settings.}
DTU is object-centric with controlled lighting and fixed intrinsics, whereas ScanNet++ is scene-centric with wider fields of view and larger viewpoint variations.
This results in a noticeable domain shift in both scene geometry and appearance statistics.

\noindent\textbf{Results.}
Table~\ref{tab:dtu} reports metrics on DTU~\cite{aanas2016dtu}. MAC-Splat outperforms PixelSplat, MVSplat, and Splatt3R on all three metrics, despite being trained only on ScanNet++. These results suggest that MAC-Splat transfers favorably under the different imaging conditions of DTU. Fig.~\ref{fig:cross_dtu} shows qualitative examples.

\subsection{Ablation Studies}
\label{sec:ablations}

We evaluate two ablated variants and the full model on ScanNet++ to isolate the contributions of the MAC loss and the semantic fusion module.
All other losses, training schedules, and hyperparameters are kept fixed.

\noindent\textbf{Variants.}
(1) \textit{w/o MAC Loss}: the MAC loss is disabled and training uses only the standard photometric reconstruction losses.
(2) \textit{w/o DINOv3}: the MAC loss is enabled, but the Residual Semantic Fusion branch is removed, so descriptors are taken from the backbone only.
(3) \textit{Ours (Full)}: the complete MAC-Splat model with both MAC and DINOv3-based semantic fusion.

Table~\ref{tab:ablation_results} reports the quantitative ablations. Removing the MAC loss (\textit{w/o MAC Loss}) severely degrades performance, yielding only 17.12\,dB on the \textbf{Very Wide} split. Applying the MAC loss without semantic fusion (\textit{w/o DINOv3}) restores the PSNR to 20.23\,dB. By outperforming PixelSplat by 4.87\,dB, this MAC-only variant indicates that explicit 3D geometric regularization, rather than DINOv3 alone, primarily drives our wide-baseline robustness.

Finally, \textit{Ours (Full)} achieves the best results across all splits. While MAC secures the global 3D geometry, DINOv3 enhances descriptor distinctiveness, further reducing LPIPS by $\sim$22\% and preserving cleaner thin structures (Fig.~\ref{fig:qualitative_ablation}).
\begin{table*}[t!]
    \centering
    \caption{Ablations on ScanNet++. We report PSNR ($\uparrow$), SSIM ($\uparrow$), and LPIPS ($\downarrow$) across the four $(\phi,\psi)$ splits. Splatt3R and MASt3R are included for reference. Best results per column are in \textbf{bold}.}
    \label{tab:ablation_results}
    \resizebox{\textwidth}{!}{%
    \begin{tabular}{l|ccc|ccc|ccc|ccc}
        \toprule
        \multirow{2}{*}{\textbf{Method}} & \multicolumn{3}{c|}{\textbf{Close}} & \multicolumn{3}{c|}{\textbf{Medium}} & \multicolumn{3}{c|}{\textbf{Wide}} & \multicolumn{3}{c}{\textbf{Very Wide}} \\
        & \multicolumn{3}{c|}{$(\phi{=}0.9,\ \psi{=}0.9)$} & \multicolumn{3}{c|}{$(\phi{=}0.7,\ \psi{=}0.7)$} & \multicolumn{3}{c|}{$(\phi{=}0.5,\ \psi{=}0.5)$} & \multicolumn{3}{c}{$(\phi{=}0.3,\ \psi{=}0.3)$} \\
        \cmidrule(lr){2-4} \cmidrule(lr){5-7} \cmidrule(lr){8-10} \cmidrule(lr){11-13}
        & PSNR & SSIM & LPIPS & PSNR & SSIM & LPIPS & PSNR & SSIM & LPIPS & PSNR & SSIM & LPIPS \\
        \midrule
        Splatt3R~\cite{lentsch2024splatt3r}              & 18.89 & 0.757 & 0.233 & 18.79 & 0.760 & 0.228 & 16.71 & 0.686 & 0.289 & 16.16 & 0.662 & 0.289 \\
        MASt3R~\cite{lentsch2024mast3r}                  & 18.44 & 0.722 & 0.292 & 15.23 & 0.648 & 0.359 & 15.72 & 0.719 & 0.295 & 15.82 & 0.608 & 0.278 \\
        \midrule
        \textit{w/o MAC Loss}                            & 19.94 & 0.791 & 0.193 & 19.45 & 0.756 & 0.203 & 20.10 & 0.780 & 0.186 & 17.12 & 0.695 & 0.278 \\
        \textit{w/o DINOv3}                              & 20.57 & 0.749 & 0.237 & 20.83 & 0.766 & 0.204 & 20.71 & 0.776 & 0.199 & 20.23 & 0.785 & 0.203 \\
        \midrule
        Ours (Full)                                      & \textbf{22.76} & \textbf{0.810} & \textbf{0.146} & \textbf{22.60} & \textbf{0.816} & \textbf{0.135} & \textbf{22.09} & \textbf{0.818} & \textbf{0.142} & \textbf{21.34} & \textbf{0.818} & \textbf{0.158} \\
        \bottomrule
    \end{tabular}
    }
     
\end{table*}

\section{Conclusion}
\label{sec:conclusion}

We presented MAC-Splat, a 3D Gaussian splatting framework for sparse-view reconstruction. The method combines an MASt3R backbone with a frozen DINOv3 encoder through a Residual Semantic Fusion module to obtain semantically informed descriptors and a sparse set of high-confidence correspondences. Based on these correspondences, we introduce the MAC loss, which enforces world-space consistency on the position, shape, and appearance of matched Gaussians. Together, these components form a unified pipeline for sparse-view 3D Gaussian splatting. This supervision improves stability under low-overlap and wide-baseline conditions. In such regimes, photometric losses often fail due to parallax and occlusions. By anchoring Gaussians with high-confidence correspondences and regularizing key 3D attributes, MAC-Splat preserves coherent geometry and appearance under large viewpoint changes. Experiments on ScanNet++ show clear gains over correspondence-based and generalizable 3DGS baselines, with particularly strong gains on lower-overlap splits, and zero-shot DTU results further demonstrate strong cross-dataset generalization. We believe that multi-attribute 3D consistency is a promising direction for robust sparse-view reconstruction and can extend to scenarios with unknown poses or dynamic scenes.

\section*{Acknowledgements}
This work was supported in part by the National Natural Science Foundation of China under Grant Nos.~62372359 and 62306233, the Xidian University Specially Funded Project for Interdisciplinary Exploration under Grant No.~TZJHF202513, and the NTU Nanyang Assistant Professorship Startup Grant under Grant No.~025661-00012. The authors thank the anonymous reviewers and area chairs for their constructive comments.

\bibliographystyle{splncs04}
\bibliography{main}
\end{document}